\documentclass[runningheads]{llncs}
\usepackage[T1]{fontenc}
\usepackage{graphicx,verbatim}
\usepackage{amsmath}
\usepackage{amssymb}
\usepackage{booktabs}
\usepackage{multirow}
\usepackage{makecell}
\usepackage{tabularx}
\usepackage{colortbl}  
\usepackage[table]{xcolor} 
\definecolor{boxblue}{rgb}{0.5686, 0.6745, 0.8784}
\providecommand{\boxbluec}[1]{\textcolor{boxblue}{#1}}
\definecolor{PDA}{rgb}{0.1176, 0.2196, 0.4196}
\providecommand{\PDAc}[1]{\textcolor{PDA}{#1}}
\definecolor{ASD}{rgb}{0.8784, 0.3255, 0.3137}
\providecommand{\ASDc}[1]{\textcolor{ASD}{#1}}

\usepackage[colorlinks=true,citecolor=blue,urlcolor=black]{hyperref}

\newcolumntype{A}{>{\centering\arraybackslash\hsize=1.5\hsize}X} 
\newcolumntype{B}{>{\centering\arraybackslash\hsize=0.65\hsize}X}
\newcolumntype{C}{>{\centering\arraybackslash\hsize=1.5\hsize}X}
\newcolumntype{D}{>{\centering\arraybackslash\hsize=0.65\hsize}X}
\newcolumntype{E}{>{\centering\arraybackslash\hsize=1.4\hsize}X}
\newcolumntype{F}{>{\centering\arraybackslash\hsize=0.8\hsize}X}

\usepackage{marvosym}

\begin{document}
\title{DepthPilot: From Controllability to Interpretability in Colonoscopy Video Generation}
\titlerunning{DepthPilot: From Controllability to Interpretability}
%
\author{Junhu Fu\inst{1,2} \and
Ke Chen\inst{3} \and
Weidong Guo\inst{1,2} \and
Shuyu Liang\inst{1,2} \and
Jie Xu\inst{1,2} \and
\mbox{Chen Ma\inst{1,2}} \and
Kehao Wang\inst{1} \and
Shengli Lin\inst{4,5} \and
Zeju Li\inst{1,2}\textsuperscript{\Letter} \\
Yuanyuan Wang\inst{1,2}\textsuperscript{\Letter} \and
Yi Guo\inst{1,2}\textsuperscript{\Letter} \and
Shuo Li\inst{6, 7}
}
\authorrunning{J. Fu et al.}
%
\institute{Princeton University, Princeton NJ 08544, USA \and
Springer Heidelberg, Tiergartenstr. 17, 69121 Heidelberg, Germany
\email{lncs@springer.com}\\
\url{http://www.springer.com/gp/computer-science/lncs} \and
ABC Institute, Rupert-Karls-University Heidelberg, Heidelberg, Germany\\
\email{\{abc,lncs\}@uni-heidelberg.de}}

\institute{College of Biomedical Engineering, Fudan University, Shanghai 200433, China \and
Key Laboratory of Medical Imaging Computing and Computer Assisted Intervention of Shanghai, Shanghai 200032, China\\
\email{\{zejuli, yywang, guoyi\}@fudan.edu.cn} \and
Department of Endoscopy, Fudan University Shanghai Cancer Center, Shanghai 200032, China \and
Endoscopy Center and Endoscopy Research Institute, Zhongshan Hospital, Fudan University, Shanghai 200032, China \and
Shanghai Collaborative Innovation Center of Endoscopy, Shanghai 200032, China \and
Department of Biomedical Engineering, Case Western Reserve University, Cleveland, OH 44106, USA \and
Department of Computer and Data Science, Case Western Reserve University, Cleveland, OH 44106, USA\\
}



\maketitle              
\begin{abstract}
Controllable medical video generation has achieved remarkable progress, but it still lacks interpretability, which requires the alignment of generated contents with physical priors and faithful clinical manifestations. To push the boundaries from mere controllability to interpretability, we propose DepthPilot, the first interpretable framework for colonoscopy video generation. This work takes a step toward trustworthy generation through two synergistic paradigms. To achieve explicit geometric grounding, DepthPilot devises a prior distribution alignment strategy, injecting depth constraints into the diffusion backbone via parameter-efficient fine-tuning to ensure anatomical fidelity. To enhance intrinsic nonlinear modeling under these geometric constraints, DepthPilot employs an adaptive spline denoising module, replacing fixed linear weights with learnable spline functions to capture complex spatio-temporal dynamics. Extensive evaluations across three public datasets and in-house clinical data confirm DepthPilot’s robust ability to produce physically consistent videos. It achieves FID scores below 15 across all benchmarks and ranks first in clinician assessments, bridging the gap between “visually realistic” and “clinically interpretable”. Moreover, DepthPilot-generated videos are expected to enable reliable 3D reconstruction, facilitating surgical navigation and blind region identification, and serve as a foundation toward the colorectal world model.


\end{abstract}
\section{Introduction} \label{Intro}
\begin{figure}[t]
\centering
\includegraphics[width=1.0\textwidth]{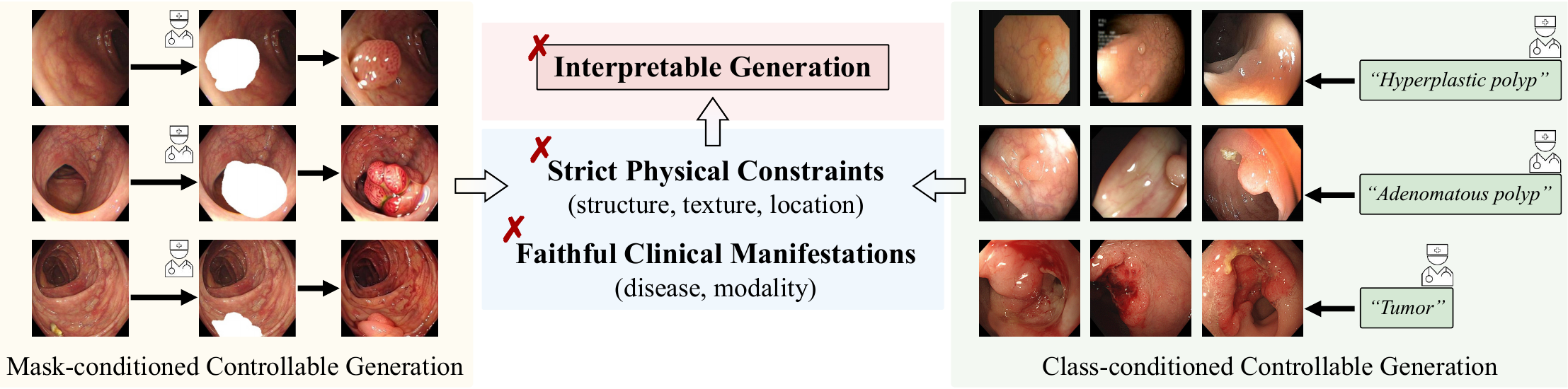}
\caption{Limitations of controllable generation methods: existing mask- and class-conditioned approaches struggle with strict physical constraints or faithful clinical manifestations, resulting in a lack of interpretability. Some images are adapted from~\cite{ControlPolypNet}.}
\label{fig1}
\end{figure}

Controllable medical video generation has emerged as a promising paradigm to alleviate the scarcity of high-quality data while providing dynamic, multi-perspective information. This paradigm enables the synthesis of lesions at specified masks~\cite{SIDM,ControlPolypNet} or the generation of data with designated classes~\cite{ColoDiff,EndoGen}. As shown in Fig.~\ref{fig1}, these algorithms represent a progress from uncontrollable to controllable for generative models, but still lack clinical interpretability. To be specific, they fail to ensure that generated contents comply with physical priors or faithfully reflect imaging characteristics. For instance, in terms of colonoscopy, hyperplastic polyps tend to display flatter shapes and lower vascular densities compared to adenomatous ones~\cite{D2PolypNet}. Another example is that polyps are more prone to occur in the descending sigmoid colon and rectum than in other regions~\cite{CRC}. Given the above situations, there is an urgent need to construct an interpretable generative model for trustworthy data synthesis, ensuring that generated colonoscopy videos are both visually realistic and physically consistent.

Recent colonoscopy video generation primarily targets content controllability and temporal coherence. For content controllability, ControlPolypNet~\cite{ControlPolypNet} and SIDM~\cite{SIDM} are both diffusion-based, adopting the idea of local generation with background reuse to synthesize lesions at specified locations. ColoDiff~\cite{ColoDiff} and EndoGen~\cite{EndoGen} construct distinct representation embeddings for different categories, thus guiding the generation of videos corresponding to various diseases, modalities, and other scenarios to improve downstream task performance. For temporal coherence, GAN-based approaches, represented by StyleGAN-V~\cite{StyleGAN-V} and MoStGAN-V~\cite{MoStGAN-V}, achieve complex feature modeling through continuous temporal encoding and motion disentanglement. Diffusion models have introduced a new paradigm with enhanced training stability and superior generation quality~\cite{DDPM,LDM,Diffusion}. Specifically, LVDM~\cite{LVDM} mitigates error accumulation through latent perturbation, while Endora~\cite{Endora} and FEAT-L~\cite{FEAT} explicitly tailor interlaced spatio-temporal attention for high-fidelity endoscopic synthesis.

While previous efforts demonstrate the feasibility of controllable colonoscopy video generation, translating controllability into interpretability remains challenging: (1) Interpretable generation, characterized by faithful intestinal structure and mucosal texture, requires strict alignment to anatomical constraints, which current controllable generative models struggle to maintain. Achieving interpretability demands explicit physical priors to provide geometric guidance, strictly anchoring the generation process to realistic anatomical spaces. (2) Once geometric grounding is introduced, generative models must handle irregular intestinal structure and rapid camera motion. Current diffusion architectures, relying heavily on linear operations like convolution, attention, and multi-layer perceptron (MLP), fail to model such complex spatio-temporal manifolds. Overcoming this bottleneck demands a backbone with superior nonlinear representation power, which effectively prevents intra-frame blur and inter-frame incoherence.

Thereby, we propose DepthPilot, the first interpretable model for colonoscopy video generation, where synthetic frames are geometrically grounded through physical prior guidance. Our contributions are three-fold: (1) Explicit geometric grounding via \textbf{Prior Distribution Alignment} (PDA). To push the boundaries from controllability to interpretability, we devise a PDA strategy that injects depth constraints into the diffusion backbone. By utilizing parameter-efficient fine-tuning, this strategy strictly enforces spatial constraints and ensures anatomical fidelity. (2) Intrinsic nonlinear modeling via \textbf{Adaptive Spline Denoising} (ASD). To capture complex spatio-temporal dynamics under geometric constraints, we introduce an ASD module that innovates the denoising architecture. By replacing fixed linear weights with learnable spline functions, this module provides superior nonlinear representation ability. (3) Extensive evaluations on three public datasets and hospital data confirm DepthPilot’s capacity to produce physically consistent colonoscopy videos. DepthPilot achieves FID scores below 15 across all datasets and ranks first in clinical assessments.

\section{Methodology}
\begin{figure}[t]
\centering
\includegraphics[width=1.0\textwidth]{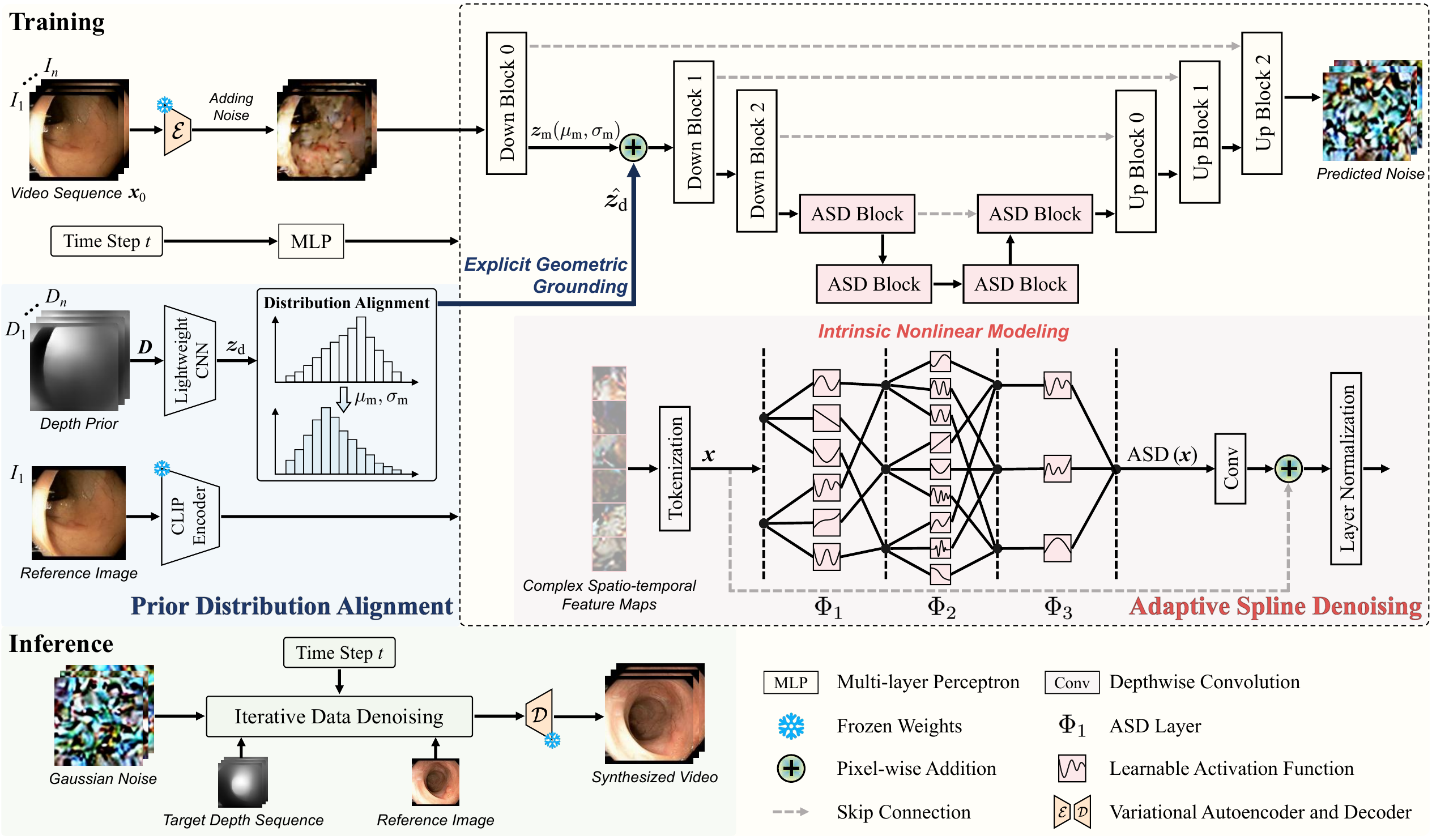}
\caption{The overall workflow of DepthPilot. The \PDAc{PDA strategy} injects geometric grounding via depth-based physical prior, while the \ASDc{ASD module} enhances nonlinear capacity to model complex spatio-temporal dynamics under such geometric constraints.}
\label{fig2}
\end{figure}

Fig.~\ref{fig2} illustrates the workflow of DepthPilot, our diffusion-based framework for interpretable colonoscopy video generation. During training, DepthPilot learns to predict the noise added in the diffusion process, establishing a geometrically grounded denoising pathway through two complementary designs. Specifically, to achieve explicit geometric grounding, our PDA strategy leverages a lightweight encoder coupled with distribution alignment to inject depth sequence priors. To capture complex spatio-temporal dynamics under these geometric constraints, our ASD module introduces adaptive spline functions into the denoising backbone, thereby enhancing nonlinear representation capacity. During inference, DepthPilot executes an iterative denoising process starting from Gaussian noise, strictly aligning with the target depth sequence to generate colonoscopy videos that are both visually realistic and clinically interpretable.

\subsection{PDA Strategy Injects Physical Prior for Interpretability}
While current generative models excel at synthesizing controllable data, they struggle to guarantee the generated contents align with physical priors, leading to a lack of clinical interpretability. To bridge this gap, we propose the PDA strategy, which explicitly injects physical priors to ensure trustworthy generation.

As shown in Fig.~\ref{fig2}, our PDA strategy leverages monocular depth sequence of the input video as primary physical prior to govern the geometric consistency of generated video, supplemented by the CLIP-encoded initial frame for style reference. To effectively embed the depth information, we employ a trainable, lightweight convolutional encoder, denoted as $\mathcal{F}_\text{l}$, to map the depth sequence $\boldsymbol{D}$ into a latent embedding $\boldsymbol{z}_\text{d} = \mathcal{F}_\text{l}(\boldsymbol{D}; \mathrm{\Theta}_\text{l})$, where $\mathcal{F}(\cdot ; \mathrm{\Theta})$ denotes a neural model with learnable parameters $\mathrm{\Theta}$.

However, directly injecting heterogeneous depth features into a pre-trained denoising backbone may disrupt the learned distribution, leading to training collapse~\cite{ControlNext}. To address this, we introduce a distribution alignment mechanism, which aligns the depth embedding $\boldsymbol{z}_\text{d}$ with feature map $\boldsymbol{z}_\text{m}$ from the main backbone. The normalized scalar $\hat{z_\text{d}}$, which constitutes tensor $\hat{\boldsymbol{z}_\text{d}}$, is formulated as
\begin{equation}
    \hat{z_\text{d}} = \frac{z_\text{d} - \mu_\text{m}}{\sqrt{\sigma_\text{m}^{2} + \epsilon}} \cdot \gamma,
    \label{eq:Alignment}
\end{equation}
where $\epsilon$ is a small constant for numerical stability, and $\gamma$ is a learnable parameter to scale the normalized value. This alignment strategy ensures the depth guidance is injected smoothly, preventing sudden disturbances to backbone features.

To ensure optimization stability and enhance representation ability, DepthPilot employs a two-stage training paradigm. During the \textbf{warm-up stage}, the model is trained unconditionally to capture the fundamental distribution of colonoscopy videos. Upon convergence, the \textbf{injection stage} activates the PDA strategy. To achieve anatomical fidelity while mitigating catastrophic forgetting, we freeze most network parameters and exclusively fine-tune the ASD blocks, leveraging their robust nonlinear modeling capabilities (Section~\ref{ASD Module}). This parameter-efficient fine-tuning process can be formally expressed as
\begin{equation}
    \boldsymbol{y} = \mathcal{F}_\text{m}\left(\boldsymbol{x}_0, \hat{\boldsymbol{z}_\text{d}}; \mathrm{\Theta}_{\text{ASD}}\right),
    \label{eq:PEFT}
\end{equation}
where $\mathcal{F}_\text{m}$ denotes diffusion backbone, $\boldsymbol{x}_0$ indicates input video, and $\mathrm{\Theta}_{\text{ASD}}$ represents trainable parameters of ASD, a subset of total model parameters $\mathrm{\Theta}_{\text{m}}$.

During inference, this geometrically grounded framework helps DepthPilot generate colonoscopy videos that adhere to structure, texture, motion, and other priors, marking a forward step from controllable to interpretable generation.

\subsection{ASD Module Empowers Nonlinear Spatio-temporal Modeling} \label{ASD Module}
While the PDA strategy successfully injects depth prior, accurately capturing complex spatio-temporal patterns under these geometric constraints remains a challenge. Current denoising backbones rely heavily on linear operations, inherently lacking the capacity to represent highly nonlinear dynamics, thus resulting in inter-frame incoherence and intra-frame blur.

To this end, DepthPilot integrates ASD module into the denoising backbone. This architecture is inspired by the Kolmogorov-Arnold representation theorem~\cite{U-KAN,KAN}, claiming that any multi-variate continuous function $f$ on a bounded domain can be represented as a finite composition of continuous single-variable functions and the operation of addition. Formally, for an input vector $\boldsymbol{v} = [v_1, v_2, \dots, v_n]$, the target function $f(\boldsymbol{v})$ can be expressed as
\begin{equation}
    f(\boldsymbol{v}) = \sum_{q=1}^{2n+1} \mathrm{\Phi}_q ( \sum_{p=1}^n \phi_{q,p}(v_p) ),
    \label{eq:KAN}
\end{equation}
where $\phi_{q,p}$ and $\mathrm{\Phi}_q$ are continuous univariate functions. Unlike current methods that learn linear weight matrices, DepthPilot parameterizes the network by learning activation functions $\phi$ on the edges connecting neurons, thereby replacing fixed activation mapping with learnable, adaptive curves.

Specifically, we parameterize these learnable univariate functions with B-splines, a set of basis functions that can be linearly combined to approximate any target function. A learnable activation function $\phi(x)$ is defined as
\begin{equation}
    \phi(x) = w_b \cdot b(x) + w_s \cdot \sum_{i} c_i B_i(x),
    \label{eq:Activation}
\end{equation}
where $b(x)$ is a SiLU function for training stability, $B_i(x)$ are the B-spline basis functions, and $c_i$ are the learnable coefficients optimized via gradient descent. Eq.~(\ref{eq:Activation}) allows DepthPilot to adaptively adjust activation function shapes during training, enhancing its nonlinear fitting capability in the representation space.

Additionally, the granularity of approximation depends on B-splines, specifically the number of grid intervals and the spline order. Increasing these parameters enables more precise fitting but introduces higher computational complexity. To balance representation power and efficiency, we employ cubic B-splines.

The integration of ASD module enables accurate noise prediction from complex spatio-temporal features, yielding high-fidelity generation that successfully captures irregular intestinal structures and motion patterns.

\subsection{DepthPilot as a Conditional Diffusion Model}
We build DepthPilot as a conditional diffusion model operating in the latent space~\cite{LDM}. To ensure computational efficiency, the input video $\boldsymbol{x}_0$ is first mapped to a latent code $\boldsymbol{z}_0$ via the encoder $\mathcal{E}$ of a pre-trained VAE, and subsequently reconstructed by the decoder $\mathcal{D}$~\cite{VAE}. The framework involves a forward process that incrementally corrupts $\boldsymbol{z}_0$ by adding Gaussian noise to obtain $\boldsymbol{z}_t$ at timestep $t$, and a learnable reverse process $p_\theta$ that reconstructs the clean data.

Specifically, the forward transition is governed by a fixed schedule $\beta_t \in (0, 1)$, expressed as $q(\boldsymbol{z}_t | \boldsymbol{z}_{t-1}) = \mathcal{N}(\boldsymbol{z}_t| \sqrt{1-\beta_t}\boldsymbol{z}_{t-1}, \beta_t \boldsymbol{I})$. The reverse process employs a denoising network $\epsilon_\theta$ to predict the noise component within $\boldsymbol{z}_t$, conditioned on the physical prior $\boldsymbol{c}$ derived from the depth sequence. The training objective of DepthPilot is to minimize the mean squared error loss:
\begin{equation}
    \mathcal{L}_{\text{DepthPilot}} = \mathbb{E}_{\boldsymbol{z}_0, \epsilon \sim \mathcal{N}(0,1), t} \big[ \| \epsilon - \epsilon_\theta(\boldsymbol{z}_t, \boldsymbol{c}, t) \|^2_2 \big],
    \label{eq:Loss}
\end{equation}
where $t$ is uniformly sampled from $\{1, \dots, T\}$ and $\epsilon$ represents Gaussian noise.

\section{Experiments}
\subsection{Experimental Settings}
\subsubsection{Datasets and Metrics.} We collect 5,136 colonoscopy video clips from three public datasets and ethics-approved hospital data. The public sources include Colonoscopic (152 original videos)~\cite{Colonoscopic}, HyperKvasir (373 original videos)~\cite{HyperKvasir}, and SUN-SEG (285 original videos)~\cite{SUN-SEG}, providing disease type annotations. The hospital database provides 203 long videos recording complete colonoscopy procedures, with lesions covering hyperplastic polyps, adenomatous polyps, and tumors. All datasets are randomly partitioned with 80\% for training and 20\% for evaluation. We acquire depth sequences using a self-supervised monocular depth estimation model~\cite{EndoDAV} tailored for endoscopy videos. Crucially, DepthPilot accepts target depth from diverse sources, encompassing real video estimate~\cite{RNNSLAM}, simulation~\cite{GaussianPancake,Simulation}, and phantom~\cite{C3VD,C3VDv2}. This broad compatibility enables interpretable generation of targeted lesions and anatomical structures.

For comparison and ablation experiments, we employ Fréchet Inception Distance (FID)~\cite{FID}, Fréchet Video Distance (FVD)~\cite{FVD}, Inception Score (IS)~\cite{IS}, and Clinician Score (CS). Specifically, FID, FVD, and IS evaluate the realism, coherence, and diversity of the synthesized contents by measuring the distributional distance between generated and real data in the feature space. We also introduce CS, where three senior clinicians rate videos on a scale of 0-5. This assessment focuses on perceptual quality and physical interpretability, such as the correspondence between polyp shapes and vascular densities, as well as the anatomical plausibility of lesion occurrences in specific regions.

\subsubsection{Implementation Details.} We implement DepthPilot in PyTorch using two NVIDIA A100 GPUs, processing 16-frame video inputs resized to 256$\times$256. The pre-trained stable video diffusion~\cite{SVD,ControlNext} serves as our generative backbone, with VAE and CLIP image encoders frozen to preserve prior knowledge~\cite{VAE,CLIP}. During training, the upper limit of iterations is set to be 500,000 with an early stop. We choose a learning rate of 1$\times10^{-4}$, a weight decay of 1$\times10^{-2}$, and the AdamW optimizer. To ensure memory efficiency and training stability, we apply mixed precision training alongside an exponential moving average strategy.

\begin{table*}[t]
\centering
\caption{Quantitative comparison with SOTA algorithms on four datasets. Best results are in \textbf{bold}.}
\label{tab:comparison}

\setlength{\tabcolsep}{1.5pt} 

\resizebox{\textwidth}{!}{
    \scriptsize 
    \begin{tabular}{lcccccccccccccccc}
    \toprule
    \multirow{2}{*}{Method} & 
    \multicolumn{4}{c}{Colonoscopic \cite{Colonoscopic}} & 
    \multicolumn{4}{c}{HyperKvasir \cite{HyperKvasir}} & 
    \multicolumn{4}{c}{SUN-SEG \cite{SUN-SEG}} & 
    \multicolumn{4}{c}{Hospital Database} \\
    \cmidrule(lr){2-5} \cmidrule(lr){6-9} \cmidrule(lr){10-13} \cmidrule(lr){14-17}
    & FVD$\downarrow$ & FID$\downarrow$ & IS$\uparrow$ & CS$\uparrow$ 
    & FVD$\downarrow$ & FID$\downarrow$ & IS$\uparrow$ & CS$\uparrow$ 
    & FVD$\downarrow$ & FID$\downarrow$ & IS$\uparrow$ & CS$\uparrow$ 
    & FVD$\downarrow$ & FID$\downarrow$ & IS$\uparrow$ & CS$\uparrow$ \\
    \midrule
    StyleGAN-V \cite{StyleGAN-V} & 2111 & 226.1 & 2.12 & 2.96 & 729 & 88.5 & 1.98 & 2.68 & 682 & 46.5 & 3.91 & 3.15 & 703 & 51.7 & 3.42 & 2.87 \\
    MoStGAN-V \cite{MoStGAN-V}   & 469  & 53.2  & 3.37 & 3.52 & 607 & 29.4 & 2.74 & 3.09 & 412 & 28.6 & 4.04 & 3.79 & 522 & 46.6 & 3.59 & 3.30 \\
    LVDM \cite{LVDM}            & 1037 & 96.9  & 1.93 & 3.20 & 528 & 49.7 & 2.15 & 2.77 & 387 & 35.9 & 3.77 & 3.42 & 416 & 40.3 & 3.12 & 3.06 \\
    Endora \cite{Endora}        & 461  & 13.4  & 3.90 & 3.47 & 496 & 19.6 & 3.29 & 2.95 & 375 & 17.8 & 4.59 & 3.86 & 389 & 23.2 & 3.65 & 3.26 \\
    FEAT-L \cite{FEAT}         & 351  & 12.3  & 4.01 & 4.09 & 511 & 21.2 & 3.27 & 3.56 & 356 & 13.6 & 4.61 & 4.23 & 402 & 18.9 & 3.88 & 3.92 \\
    ColoDiff \cite{ColoDiff}    & 339  & 12.7  & 3.95 & 4.26 & 473 & 16.3 & 3.46 & 3.76 & 294 & 11.9 & 4.78 & 4.55 & 336 & 15.7 & 4.14 & 3.88 \\
    
    \rowcolor{gray!15} 
    DepthPilot (Ours)           & \textbf{289} & \textbf{11.8} & \textbf{4.36} & \textbf{4.58} & \textbf{392} & \textbf{14.2} & \textbf{3.58} & \textbf{3.78} & \textbf{272} & \textbf{10.3} & \textbf{4.96} & \textbf{4.71} & \textbf{284} & \textbf{13.7} & \textbf{4.32} & \textbf{4.05} \\
    \bottomrule
    \end{tabular}
}
\end{table*}

\subsection{Comparison with SOTA Methods}
Table~\ref{tab:comparison} compares DepthPilot with six SOTA baselines across four datasets. DepthPilot demonstrates superior performance, consistently outperforming GAN-based methods and pushing the boundary of diffusion-based ones. It achieves an FID below 15 across all datasets, indicating exceptional image quality that closely approximates real data distribution. Notably, it attains a 272 FVD and a 4.71 CS on the SUN-SEG dataset~\cite{SUN-SEG}, indicating the generated videos possess high coherence and realism, effectively “passing for real” in clinical assessments. The results indicate that depth prior guidance effectively anchors generated dynamics to both physical laws and perceptual constraints.

\begin{figure}[t]
\centering
\includegraphics[width=1.0\textwidth]{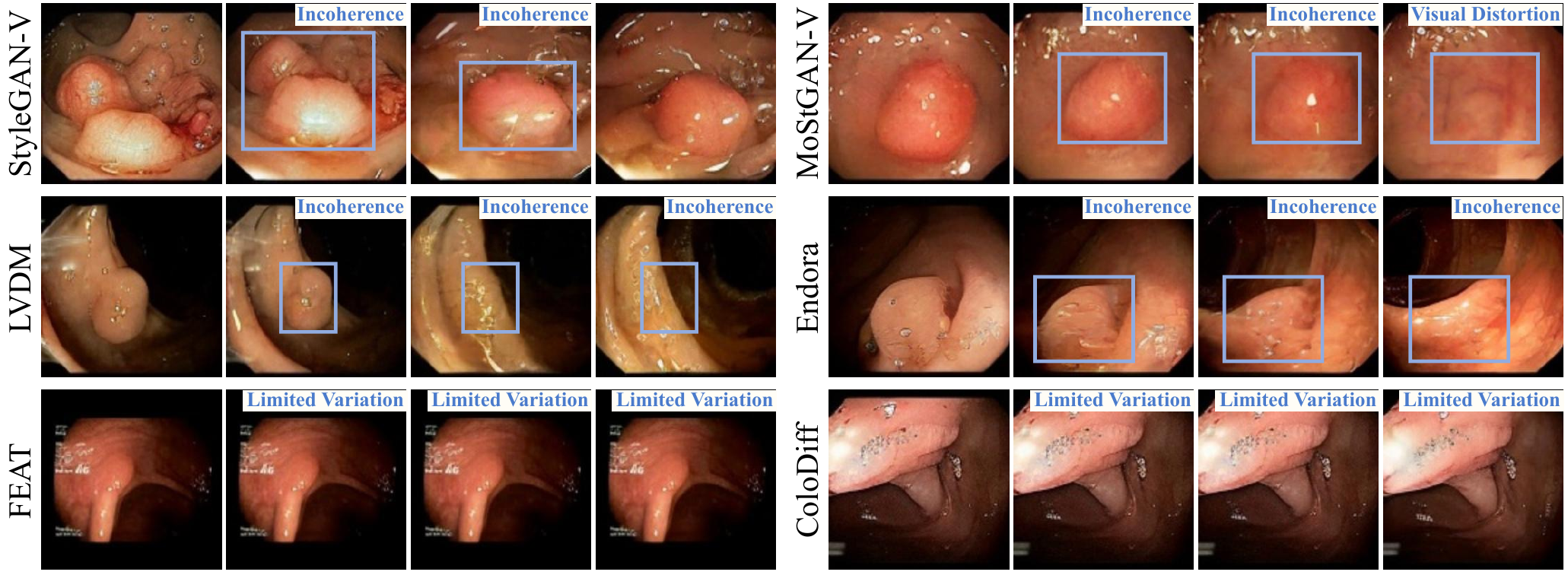}
\caption{Visual comparison of generated videos by other compared methods. The \boxbluec{blue} boxes indicate regions with corresponding issues, including inter-frame incoherence, limited content variation, and anatomical or textural visual distortion.}
\label{fig3}
\end{figure}

\begin{figure}[t]
\centering
\includegraphics[width=1.0\textwidth]{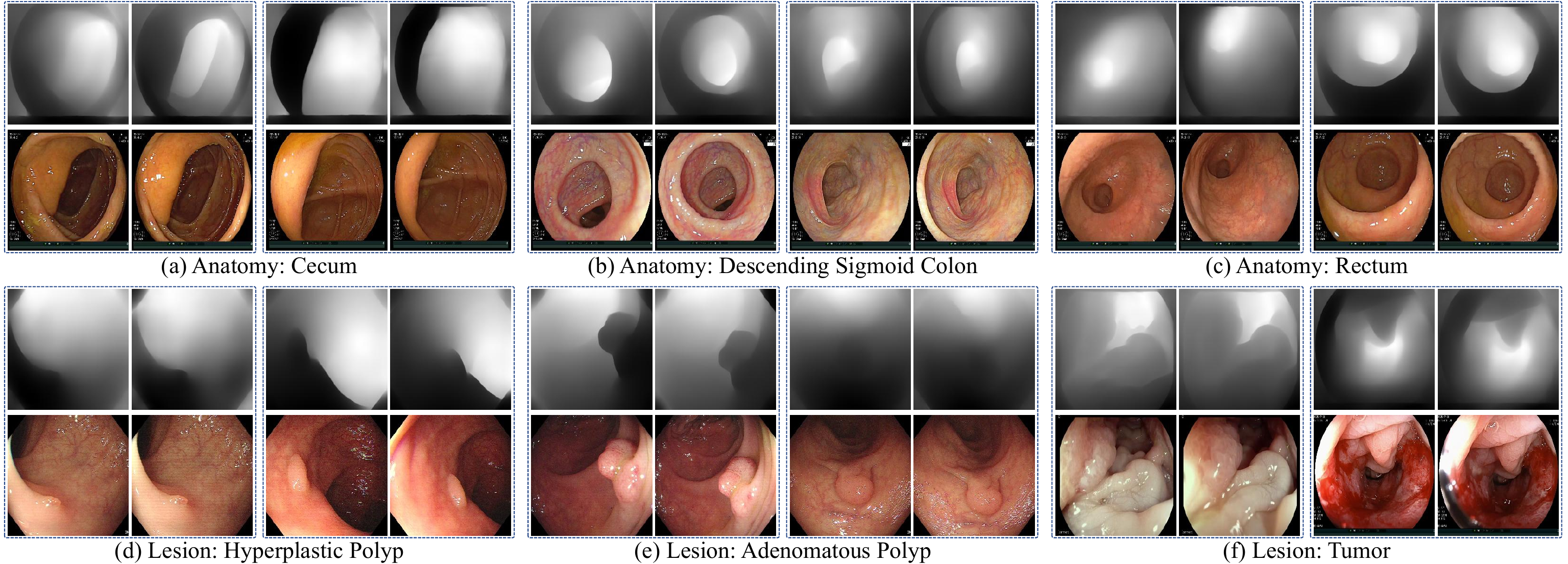}
\caption{Examples of videos generated by DepthPilot under depth prior guidance. (a)-(c) demonstrate specific anatomical structures, and (d)-(f) demonstrate specific lesions.}
\label{fig4}
\end{figure}

Visualizations in Fig.~\ref{fig3} corroborate our quantitative finding. Without explicit physical constraints or robust temporal modeling, the compared methods exhibit inter-frame incoherence, limited content variation, and anatomical distortion. In contrast, Fig.~\ref{fig4} shows DepthPilot generating anatomically faithful videos across key regions: cecum (prone to polyp miss), descending sigmoid colon, and rectum (high-risk sites). Notably, DepthPilot precisely matches target depth topology, rendering flatter hyperplastic polyps and denser vascular adenomatous polyps. This structural and textural fidelity stems from physical priors via PDA strategy and nonlinear manifold capture via ASD module. Additional generated videos are provided in the \textbf{Supplementary Material}. As noted before, beyond \textit{in-vivo} depth estimate~\cite{RNNSLAM}, DepthPilot is also compatible with depth priors from simulation~\cite{GaussianPancake,Simulation} and phantom~\cite{C3VD,C3VDv2}, unlocking broad applications.

\subsection{Ablation Experiments}

\begin{figure}[t]
    \centering
    \begin{minipage}[t]{0.48\textwidth}
        \vspace{0pt} 
        
        \makeatletter\def\@captype{table}\makeatother 
        
        \caption{Ablation results for PDA strategy and ASD module. Best results are in \textbf{bold}.}
        \label{tab:ablation}

        \vspace{1.1em} 
        \scriptsize
        \renewcommand{\arraystretch}{1.2}
        \begin{tabularx}{\linewidth}{
            >{\centering\arraybackslash\hsize=1.8\hsize}X
            *{4}{>{\centering\arraybackslash\hsize=0.8\hsize}X}
        }
        \toprule
        \multirow{2}{*}{\makecell{Network\\Architecture}} & 
        \multicolumn{4}{c}{SUN-SEG \cite{SUN-SEG}} \\
        \cmidrule(lr){2-5}
        & FVD$\downarrow$ & FID$\downarrow$ & IS$\uparrow$ & CS$\uparrow$ \\
        \midrule
        Baseline & 483 & 35.6 & 3.36 & 2.91 \\
        + PDA & 347 & 15.5 & 3.82 & 3.85 \\
        \rowcolor{gray!15} 
        + PDA + ASD & \textbf{272} & \textbf{10.3} & \textbf{4.96} & \textbf{4.71} \\
        \bottomrule
        \end{tabularx}
        \renewcommand{\arraystretch}{1.0}
    \end{minipage}%
    \hfill
    \begin{minipage}[t]{0.48\textwidth}
        \vspace{0pt} 
        \centering
        
        \includegraphics[width=\linewidth]{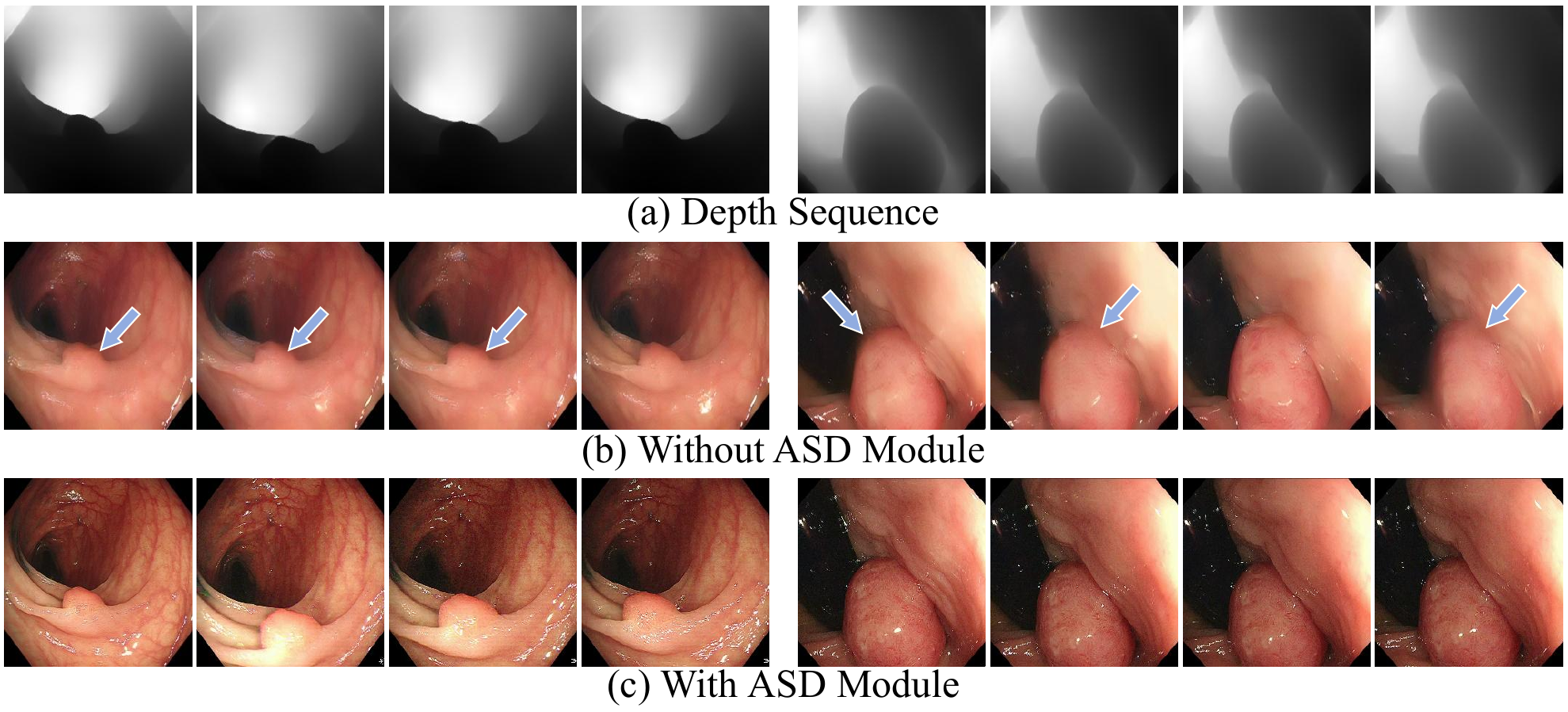}
        
        \makeatletter\def\@captype{figure}\makeatother 

        \vspace{-0.5em} 
        \caption{The visualization of ablation experiments regarding ASD module. The \boxbluec{blue} arrows indicate blurred regions.}
        \label{fig5}
    \end{minipage}
\end{figure}



Table~\ref{tab:ablation} presents the ablation study of PDA and ASD modules on SUN-SEG~\cite{SUN-SEG}. Integrating PDA strategy reduces FVD from 483 to 347, and ASD module further optimizes it to 272. As shown in Fig.~\ref{fig5}(b), the removal of ASD leads to local blurring and structural degradation. Unlike fixed activation functions that constrain the network’s capacity to model irregular and scale-varying intestinal anatomies, ASD’s learnable activation functions equip DepthPilot to effectively capture complex spatio-temporal representation.

\section{Conclusion}
In this paper, we propose DepthPilot, a diffusion-based framework for interpretable colonoscopy video generation. The PDA strategy ensures synthesized videos follow realistic motion patterns while preserving physical properties. The ASD module enhances nonlinear representation ability, which is especially beneficial for modeling complex spatio-temporal manifolds. Comprehensive experiments on four datasets validate DepthPilot’s capacity for generating interpretable colonoscopy videos. As a pioneering step toward trustworthy generation, DepthPilot also paves the way for reliable reconstruction of intestinal structure, facilitating surgical navigation and blind region identification, and lays a solid foundation for a unified colorectal world model.

\bibliographystyle{splncs04}
\bibliography{References}

\begin{thebibliography}{10}
\providecommand{\url}[1]{\texttt{#1}}
\providecommand{\urlprefix}{URL }
\providecommand{\doi}[1]{https://doi.org/#1}

\bibitem{SVD}
Blattmann, A., Dockhorn, T., Kulal, S., Mendelevitch, D., Kilian, M., Lorenz, D., et~al.: Stable video diffusion: Scaling latent video diffusion models to large datasets. arXiv preprint arXiv:2311.15127  (2023)

\bibitem{C3VD}
Bobrow, T.L., Golhar, M., Vijayan, R., Akshintala, V.S., Garcia, J.R., Durr, N.J.: Colonoscopy 3d video dataset with paired depth from 2d-3d registration. Medical image analysis  \textbf{90},  102956 (2023)

\bibitem{GaussianPancake}
Bonilla, S., Zhang, S., Psychogyios, D., Stoyanov, D., Vasconcelos, F., Bano, S.: Gaussian pancakes: geometrically-regularized 3d gaussian splatting for realistic endoscopic reconstruction. In: International Conference on Medical Image Computing and Computer-Assisted Intervention. pp. 274--283. Springer (2024)

\bibitem{HyperKvasir}
Borgli, H., Thambawita, V., Smedsrud, P.H., Hicks, S., Jha, D., Eskeland, S.L., et~al.: Hyperkvasir, a comprehensive multi-class image and video dataset for gastrointestinal endoscopy. Scientific data  \textbf{7}(1), ~283 (2020)

\bibitem{CRC}
De~Leon, M.P., Di~Gregorio, C.: Pathology of colorectal cancer. Digestive and Liver Disease  \textbf{33}(4),  372--388 (2001)

\bibitem{D2PolypNet}
Fu, J., Gao, Y., Zhou, P., Huang, Y., Jiao, J., Lin, S., et~al.: D2polyp-net: A cross-modal space-guided network for real-time colorectal polyp detection and diagnosis. Biomedical Signal Processing and Control  \textbf{91},  105934 (2024)

\bibitem{ColoDiff}
Fu, J., Liang, S., Li, W., Ma, C., Huang, P., Wang, K., et~al.: Colodiff: Integrating dynamic consistency with content awareness for colonoscopy video generation. arXiv preprint arXiv:2602.23203  (2026)

\bibitem{C3VDv2}
Golhar, M.V., Fretes, L.S.G., Ayers, L., Akshintala, V.S., Bobrow, T.L., Durr, N.J.: C3vdv2--colonoscopy 3d video dataset with enhanced realism. arXiv preprint arXiv:2506.24074  (2025)

\bibitem{LVDM}
He, Y., Yang, T., Zhang, Y., Shan, Y., Chen, Q.: Latent video diffusion models for high-fidelity long video generation. arXiv preprint arXiv:2211.13221  (2022)

\bibitem{SIDM}
Heo, C., Jung, J.: Semantic interpolative diffusion model: Bridging the interpolation to masks and colonoscopy image synthesis for robust generalization. In: International Conference on Medical Image Computing and Computer-Assisted Intervention. pp. 519--529. Springer (2025)

\bibitem{FID}
Heusel, M., Ramsauer, H., Unterthiner, T., Nessler, B., Hochreiter, S.: Gans trained by a two time-scale update rule converge to a local nash equilibrium. Advances in neural information processing systems  \textbf{30} (2017)

\bibitem{DDPM}
Ho, J., Jain, A., Abbeel, P.: Denoising diffusion probabilistic models. Advances in neural information processing systems  \textbf{33},  6840--6851 (2020)

\bibitem{SUN-SEG}
Ji, G.P., Xiao, G., Chou, Y.C., Fan, D.P., Zhao, K., Chen, G., et~al.: Video polyp segmentation: A deep learning perspective. Machine Intelligence Research  \textbf{19}(6),  531--549 (2022)

\bibitem{VAE}
Kingma, D.P., Welling, M.: Auto-encoding variational bayes. arXiv preprint arXiv:1312.6114  (2013)

\bibitem{Endora}
Li, C., Liu, H., Liu, Y., Feng, B.Y., Li, W., Liu, X., et~al.: Endora: Video generation models as endoscopy simulators. In: International conference on medical image computing and computer-assisted intervention. pp. 230--240. Springer (2024)

\bibitem{U-KAN}
Li, C., Liu, X., Li, W., Wang, C., Liu, H., Liu, Y., et~al.: U-kan makes strong backbone for medical image segmentation and generation. In: Proceedings of the AAAI conference on artificial intelligence. vol.~39, pp. 4652--4660 (2025)

\bibitem{EndoGen}
Liu, X., Liu, H., Wang, C., Liu, T., Yuan, Y.: Endogen: Conditional autoregressive endoscopic video generation. In: International Conference on Medical Image Computing and Computer-Assisted Intervention. pp. 169--179. Springer (2025)

\bibitem{KAN}
Liu, Z., Wang, Y., Vaidya, S., Ruehle, F., Halverson, J., Solja{\v{c}}i{\'c}, M., et~al.: Kan: Kolmogorov-arnold networks. arXiv preprint arXiv:2404.19756  (2024)

\bibitem{RNNSLAM}
Ma, R., Wang, R., Zhang, Y., Pizer, S., McGill, S.K., Rosenman, J., et~al.: Rnnslam: Reconstructing the 3d colon to visualize missing regions during a colonoscopy. Medical image analysis  \textbf{72},  102100 (2021)

\bibitem{Colonoscopic}
Mesejo, P., Pizarro, D., Abergel, A., Rouquette, O., Beorchia, S., Poincloux, L., et~al.: Computer-aided classification of gastrointestinal lesions in regular colonoscopy. IEEE transactions on medical imaging  \textbf{35}(9),  2051--2063 (2016)

\bibitem{ControlNext}
Peng, B., Wang, J., Zhang, Y., Li, W., Yang, M.C., Jia, J.: Controlnext: Powerful and efficient control for image and video generation. arXiv preprint arXiv:2408.06070  (2024)

\bibitem{CLIP}
Radford, A., Kim, J.W., Hallacy, C., Ramesh, A., Goh, G., Agarwal, S., et~al.: Learning transferable visual models from natural language supervision. In: International conference on machine learning. pp. 8748--8763. PmLR (2021)

\bibitem{LDM}
Rombach, R., Blattmann, A., Lorenz, D., Esser, P., Ommer, B.: High-resolution image synthesis with latent diffusion models. In: Proceedings of the IEEE/CVF conference on computer vision and pattern recognition. pp. 10684--10695 (2022)

\bibitem{IS}
Salimans, T., Goodfellow, I., Zaremba, W., Cheung, V., Radford, A., Chen, X.: Improved techniques for training gans. Advances in neural information processing systems  \textbf{29} (2016)

\bibitem{ControlPolypNet}
Sharma, V., Kumar, A., Jha, D., Bhuyan, M.K., Das, P.K., Bagci, U.: Controlpolypnet: towards controlled colon polyp synthesis for improved polyp segmentation. In: Proceedings of the IEEE/CVF Conference on Computer Vision and Pattern Recognition. pp. 2325--2334 (2024)

\bibitem{MoStGAN-V}
Shen, X., Li, X., Elhoseiny, M.: Mostgan-v: Video generation with temporal motion styles. In: Proceedings of the IEEE/CVF Conference on Computer Vision and Pattern Recognition. pp. 5652--5661 (2023)

\bibitem{StyleGAN-V}
Skorokhodov, I., Tulyakov, S., Elhoseiny, M.: Stylegan-v: A continuous video generator with the price, image quality and perks of stylegan2. In: Proceedings of the IEEE/CVF conference on computer vision and pattern recognition. pp. 3626--3636 (2022)

\bibitem{Diffusion}
Sohl-Dickstein, J., Weiss, E., Maheswaranathan, N., Ganguli, S.: Deep unsupervised learning using nonequilibrium thermodynamics. In: International conference on machine learning. pp. 2256--2265. pmlr (2015)

\bibitem{FVD}
Unterthiner, T., Van~Steenkiste, S., Kurach, K., Marinier, R., Michalski, M., Gelly, S.: Towards accurate generative models of video: A new metric \& challenges. arXiv preprint arXiv:1812.01717  (2018)

\bibitem{FEAT}
Wang, H., Yang, Z., Zhang, H., Zhao, D., Wei, B., Xu, Y.: Feat: Full-dimensional efficient attention transformer for medical video generation. In: International Conference on Medical Image Computing and Computer-Assisted Intervention. pp. 267--277. Springer (2025)

\bibitem{Simulation}
Zhang, S., Zhao, L., Huang, S., Ye, M., Hao, Q.: A template-based 3d reconstruction of colon structures and textures from stereo colonoscopic images. IEEE Transactions on Medical Robotics and Bionics  \textbf{3}(1),  85--95 (2020)

\bibitem{EndoDAV}
Zhou, Z., Yang, C., Yang, P., Yang, X., Shen, W.: Endodav: Depth any video in endoscopy with spatiotemporal accuracy. In: International Conference on Medical Image Computing and Computer-Assisted Intervention. pp. 192--201. Springer (2025)

\end{thebibliography}
%






\end{document}